%% file: eccv2018submission.tex
\documentclass[runningheads]{llncs}
\usepackage{graphicx}
\usepackage{amsmath,amssymb} 
\usepackage{color}
\usepackage{tabularx}
\usepackage{wrapfig,lipsum,booktabs}

\begin{document}
\pagestyle{headings}
\mainmatter
\def\ECCV18SubNumber{2156}  

\title{Multi-modal Cycle-consistent Generalized Zero-Shot Learning}

\titlerunning{Multi-modal Cycle-consistent Generalized Zero-Shot Learning}

\authorrunning{Rafael Felix, Vijay Kumar B G, Ian Reid, Gustavo Carneiro}

\author{Rafael Felix, Vijay Kumar B G, Ian Reid, Gustavo Carneiro}
\institute{Australian Institute for Machine Learning, University of Adelaide, Australia\\
	\email{ \{rafael.felixalves,vijay.kumar,ian.reid,gustavo.carneiro\}@adelaide.edu.au}}

\maketitle

\begin{abstract}
In generalized zero shot learning (GZSL), the set of classes are split into seen and unseen classes, where training relies on the semantic features of the seen and unseen classes and the visual representations of only the seen classes, while testing uses the visual representations of the seen and unseen classes.  Current methods address GZSL by learning a transformation from the visual to the semantic space, exploring the assumption that the distribution of classes in the semantic and visual spaces is relatively similar.  Such methods tend to transform unseen testing visual representations into one of the seen classes' semantic features instead of the semantic features of the correct unseen class, resulting in low accuracy GZSL classification.  Recently, generative adversarial networks (GAN) have been explored to synthesize visual representations of the unseen classes from their semantic features - the synthesized representations of the seen and unseen classes are then used to train the GZSL classifier.  This approach has been shown to boost GZSL classification accuracy, but there is one important missing constraint: there is no guarantee that synthetic visual representations can generate back their semantic feature in a multi-modal cycle-consistent manner.  This missing constraint can result in synthetic visual representations that do not represent well their semantic features, which means that the use of this constraint can improve GAN-based approaches. In this paper, we propose the use of such constraint based on a new regularization for the GAN training that forces the generated visual features to reconstruct their original semantic features. Once our model is trained with this multi-modal cycle-consistent semantic compatibility, we can then synthesize more representative visual representations for the seen and, more importantly, for the unseen classes.  Our proposed approach shows the best GZSL classification results in the field in several publicly available datasets.
\keywords{generalized zero-shot learning; generative adversarial networks; cycle consistency loss}
\end{abstract}

\section{Introduction}

\begin{figure}[h]
\centering
\includegraphics[width=0.6\textwidth]{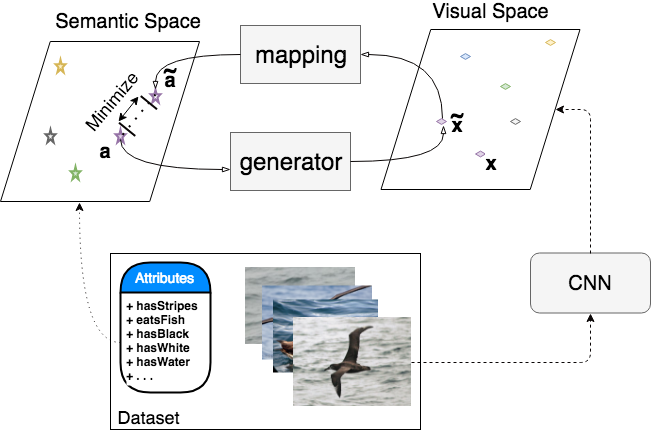}
\caption{Overview of the proposed multi-modal cycle-consistent GZSL approach. Our approach extends the idea of synthesizing  visual representations of seen and unseen classes in order to train a classifier for the GZSL problem~\cite{XianCVPR2018}.  The main contribution of the paper is the use of a new multi-modal cycle consistency loss in the training of the visual feature generator that 
minimizes the reconstruction error between the semantic feature $\mathbf{a}$, which was used to synthesize the visual feature $\widetilde{\mathbf{x}}$, and the reconstructed semantic feature $\widetilde{\mathbf{a}}$ mapped from $\widetilde{\mathbf{x}}$.  This loss is shown to constrain the optimization problem more effectively in order to produce useful synthesized visual features for training the GZSL classifier.}
\label{eq:cycle_teaser}

\end{figure}

\let\thefootnote\relax\footnotetext{All authors gratefully acknowledge the support of the Australian Research Council through the Centre of Excellence for Robotic Vision (project number CE140100016), Laureate Fellowship FL130100102 to IR and Discover Project DP180103232 to GC.}

Generalized Zero-shot Learning (GZSL) separates the classes of interest into a sub-set of seen classes and another sub-set of unseen classes.  The training process uses the semantic features of both sub-sets and the visual representations of only the seen classes; while the testing process aims to classify the visual representations of both sub-sets~\cite{XianCVPR2017,zhang2015zero}. The semantic features available for both the training and testing classes are typically acquired from other domains, such as visual features~\cite{lampert2014attribute}, text~\cite{qiao2016less,socher2013zero,zhang2015zero}, or learned classifiers~\cite{yu2013designing}. The traditional approach to address this challenge~\cite{XianCVPR2017} involves the learning of a transformation from the visual to the semantic space of the seen classes.  Testing is then performed by transforming the visual representation of the seen and unseen classes into this semantic space, where classification is typically achieved with a nearest neighbor classifier that selects the closest class in the semantic space. In contrast to Zero-shot Learning (ZSL), which uses only the unseen domain for testing, GZSL approaches tend to be biased towards the seen classes, producing poor classification results, particularly for the unseen testing classes~\cite{XianCVPR2018}. 

These traditional approaches rely on the assumption that the distributions observed in the semantic and visual spaces are relatively similar.  Recently, this assumption has been relaxed to allow the semantic space to be optimized together with the transformation from the visual to the semantic space ~\cite{long2017zero} - this alleviates the classification bias mentioned above to a certain degree.  More recent approaches consist of building a generative adversarial network (GAN) that synthesizes visual representations of the seen and unseen classes directly from their semantic representation~\cite{bucher2017generating,long2017zero}.  These synthesized features are then used to train a multi-class classifier of seen and unseen classes.  This approach has been shown to improve the GZSL classification accuracy, but an obvious weakness is that the unconstrained nature of the generation process may let the approach generate unrepresentative synthetic visual representations, particularly of the unseen classes (i.e., representations that are far from possible visual representations of the test classes).

The {\bf main contribution} of this paper is a {\bf new regularization of the generation of synthetic visual representations in the training of GAN-based methods that address the GZSL classification problem}.  This regularization is \textbf{based on a multi-modal cycle consistency loss term that enforces good reconstruction from the synthetic visual representations back to their original semantic features}
(see Fig.~\ref{eq:cycle_teaser}). This regularization is motivated by the cycle consistency loss applied in training GANs~\cite{CycleGAN2017} that forces the generative training approach to produce more constrained visual representations. We argue that this constraint preserves the semantic compatibility between visual features and semantic features.
Once our model is trained with this multi-modal cycle consistency loss term, we can then synthesize visual representations for unseen classes in order to train a GZSL classifier~\cite{tran2017bayesian,XianCVPR2018}.

Using the experimental setup described by Xian et al.~\cite{XianCVPR2018}, we show that our proposed regularization provides significant improvements not only in terms of GZSL classification accuracy, but also  
ZSL on the following datasets: Caltech-UCSD-Birds 200-2011 (CUB)~\cite{welinder2010caltech,XianCVPR2017}, Oxford-Flowers (FLO)~\cite{nilsback2008automated}, Scene Categorization Benchmark (SUN) ~\cite{farhadi2009describing,XianCVPR2017}, Animals with features (AWA)~\cite{lampert2014attribute,XianCVPR2017}, and \textit{ImageNet}~\cite{deng2009imagenet} . In fact, the experiments show that our proposed approach holds the current best ZSL and GZSL classification results in the field for these datasets.

%
%
%
%
%
%
%

\section{Literature Review}
\label{sec:lit_review}

The starting point for our literature review is the work by Xian et al.~\cite{XianCVPR2017,XianCVPR2018}, who proposed new benchmarks using commonly accepted evaluation protocols on publicly available datasets.  These benchmarks allow a fair comparison among recently proposed ZSL and GZSL approaches, and for this reason we explore those benchmarks to compare our results with the ones obtained from the current state of the art in the field.  We provide a general summary of the methods presented in~\cite{XianCVPR2017}, and encourage the reader to study that paper in order to obtain more details on previous works.  
The majority of the ZSL and GZSL methods tend to compensate the lack of visual representation of the unseen classes with the learning of a mapping between visual and semantic spaces \cite{chen2018preserving}, \cite{annadani2018relations}.
For instance, a fairly successful approach is based on a bi-linear compatibility function that associates visual representation and semantic features.  Examples of such approaches are ALE~\cite{akata2016label}, DEVISE~\cite{frome2013devise},  SJE~\cite{akata2015evaluation}, ESZSL~\cite{romera2015embarrassingly}, and SAE~\cite{kodirov2017semantic}.  Despite their simplicity, these methods tend to produce the current state-of-the-art results on benchmark datasets~\cite{XianCVPR2017}.  A straightforward extension of the methods above is the exploration of a non-linear compatibility function between visual and semantic spaces.  These approaches, exemplified by LATEM~\cite{xian2016latent} and CMT~\cite{socher2013zero}, tend not to be as competitive as their bi-linear counterpart, probably because the more complex models need larger training sets to generalize more effectively.
Seminal ZSL and GZSL methods were based on models relying on learning intermediate feature classifiers, which are combined to predict image classes (e.g., DAP and IAP)~\cite{lampert2014attribute} -- these models tend to present relatively poor classification results.
Finally, hybrid models, such as SSE~\cite{zhang2015zero}, CONSE~\cite{norouzi2013zero}, SYNC~\cite{changpinyo2016synthesized}, rely on a mixture model of seen classes to represent images and semantic embeddings.  These methods tend to be competitive for classifying the seen classes, but not for the unseen classes.

The main disadvantage of the methods above is that the lack of visual training data for the unseen classes biases the mapping between visual and semantic spaces towards the semantic features of seen classes, particularly for unseen test images.  This is an issue for GZSL because it has a negative effect in the classification accuracy of the unseen classes.  Recent research address this issue using GAN models that are trained to synthesize visual representations for the seen and unseen classes, which can then be used to train a classifier for both the seen and unseen classes~\cite{long2017zero,bucher2017generating}. However, the unconstrained generation of synthetic visual representations for the unseen classes allows the production of synthetic samples that may be too far from the actual distribution of visual representations, particularly for the unseen classes.  
In GAN literature, this problem is known as unpaired training~\cite{CycleGAN2017}, where not all source samples (e.g., semantic features) have corresponding target samples (e.g., visual features) for training.  This creates a highly unconstrained optimization problem that has been solved by Zhu et al.~\cite{CycleGAN2017} with a cycle consistency loss to push the representation from the target domain back to the source domain, which helped constraining the optimization problem.  In this paper, we explore this idea for GZSL, which is a novelty compared to previous GAN-based methods proposed in GZSL and ZSL.  

%
%
%
%
%
%
%

\section{Multi-modal Cycle-consistent Generalized Zero Shot Learning}
\label{sec:method}

In GZSL and ZSL~\cite{XianCVPR2017}, the dataset is denoted by $\mathcal{D} = \{(\textbf{x},\mathbf{a},y)_i\}_{i=1}^{|\mathcal{D}|}$ with $\textbf{x} \in \mathcal{X} \subseteq \mathbb{R}^K$ representing visual representation (e.g., image features from deep residual nets~\cite{he2016resnet}), $\mathbf{a} \in \mathcal{A} \subseteq \mathbb R^L$ denoting $L$-dimensional semantic feature (e.g., set of binary attributes~\cite{lampert2014attribute} or a dense \textit{word2vec} representation~\cite{mikolov2013distributed}), $y \in \mathcal{Y} = \{ 1,..., C \}$ denoting the image class, and $|.|$ representing set cardinality.
The set $\mathcal{Y}$ is split into seen and unseen subsets, where the seen subset is denoted by $\mathcal{Y}_S$ and the unseen subset by $\mathcal{Y}_U$, with $\mathcal{Y} = \mathcal{Y}_S \cup \mathcal{Y}_U$ and $\mathcal{Y}_S \cap \mathcal{Y}_U = \emptyset$.  The dataset $\mathcal{D}$ is also divided into mutually exclusive training and testing subsets: $\mathcal{D}^{Tr}$ and $\mathcal{D}^{Te}$, respectively.
Furthermore, the training and testing sets can also be divided in terms of the seen and unseen classes, so this means that $\mathcal{D}^{Tr}_S$ denotes the training samples of the seen classes, while $\mathcal{D}^{Tr}_U$ represents the training samples of the unseen classes (similarly for $\mathcal{D}^{Te}_S$ and $\mathcal{D}^{Te}_U$ for the testing set).
During training, samples in $\mathcal{D}_S^{Tr}$ contain the visual representation $\mathbf{x}_i$, semantic feature $\mathbf{a}_i$ and class label $y_i$; while the samples in $\mathcal{D}_U^{Tr}$ comprise only the semantic feature and class label.  During ZSL testing, only the samples from $\mathcal{D}_U^{Te}$ are used; while in GZSL testing, all samples from $\mathcal{D}^{Te}$ are used.  Note that for ZSL and GZSL problems, only the visual representation of the testing samples is used to predict the class label.

Below, we first explain the f-CLSWGAN model~\cite{XianCVPR2018}, which is the baseline for the implementation of the main contribution of this paper: the multi-modal cycle consistency loss used in the training for the feature generator in GZSL models based on GANs.  The loss, feature generator, learning and testing procedures are explained subsequently.

\begin{figure}[h]
\centering
\includegraphics[width=0.6\textwidth]{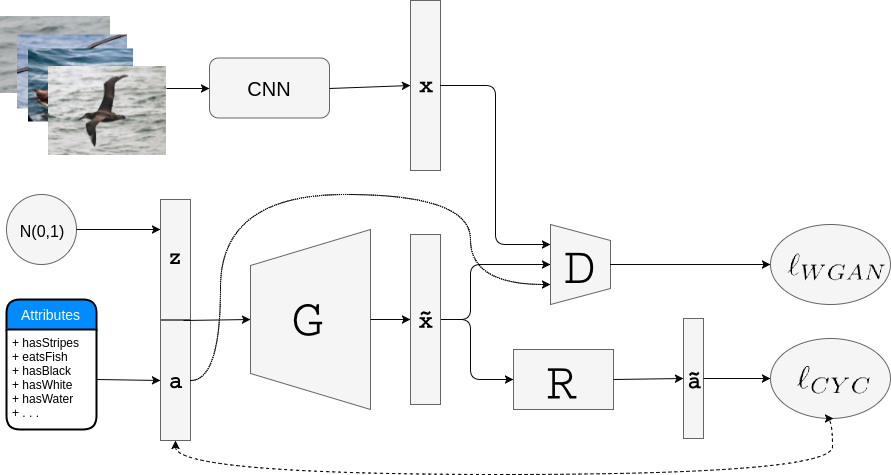}
\caption{Overview of the multi-modal cycle-consistent GZSL model. 
The visual features, represented by $\mathbf{x}$, are extracted from a state-of-art CNN model, and the semantic features, represented by $\mathbf{a}$, are available from the training set.  The generator $G(.)$ synthesizes new visual features $\widetilde{\mathbf{x}}$ using the semantic feature and a randomly sampled noise vector $\mathbf{z} \sim \mathcal{N}(\mathbf{0},\mathbf{I})$, and the discriminator $D(.)$ tries to distinguish between real and synthesized visual features.  Our main contribution is focused on the integration of a multi-modal cycle consistency loss (at the bottom) that minimizes the error between the original semantic feature $\mathbf{a}$ and its reconstruction $\widetilde{\mathbf{a}}$, produced by the regressor $R(.)$.}
\label{fig:overview}

\end{figure}

\subsection{f-CLSWGAN}
\label{sec:CLSWGAN}

Our approach is an extension of the feature generation method proposed by Xian et al.~\cite{XianCVPR2018}, which consists of a classification regularized generative adversarial network (f-CLSWGAN). This network is composed of a generative model $G:\mathcal{A} \times \mathcal{Z} \rightarrow \mathcal{X}$ (parameterized by $\theta_G$) that produces a visual representation $\widetilde{\mathbf{x}}$ given its semantic feature $\mathbf{a}$ and a noise vector $\mathbf{z} \sim \mathcal{N}(\mathbf{0},\mathbf{I})$ sampled from a multi-dimensional centered Gaussian, and a discriminative model $D:\mathcal{X} \times \mathcal{A} \rightarrow [0,1]$ (parameterized by $\theta_D$) that tries to distinguish whether the input $\mathbf{x}$ and its semantic representation $\mathbf{a}$ represent a true or generated visual representation and respective semantic feature.  
Note that while the method developed by Yan et al.~\cite{yan2016attribute2image} concerns the generation of realistic images, our proposed approach, similarly to~\cite{XianCVPR2018,long2017zero,bucher2017generating}, aims to generate visual representations, such as the features from a deep residual network~\cite{he2016resnet} - the strategy based on visual representation has shown to produce more accurate GZSL classification results compared to the use of realistic images.  
The training algorithm for estimating $\theta_G$ and $\theta_D$ follows a minimax game, where $G(.)$ generates synthetic visual representations that are supposed to fool the discriminator, which in turn tries to distinguish the real from the synthetic visual representations.  We rely on one of the most stable training methods for GANs, called Wasserstein GAN, which uses the following loss function~\cite{arjovsky2017wasserstein}:

\begin{equation}
\theta_G^*,\theta_D^*=\arg\min_{\theta_G} \max_{\theta_D} \ell_{WGAN}(\theta_G,\theta_D),  
\label{eq:wgan}
\end{equation}
with 
\begin{equation}
\begin{split}
\ell_{WGAN}(\theta_G,\theta_D) & = \mathbb E_{(\mathbf{x},\mathbf{a}) \sim \mathbb P^{x,a}}[D(\mathbf{x},\mathbf{a};\theta_D)] - \mathbb E_{(\widetilde{\mathbf{x}},\mathbf{a}) \sim \mathbb P^{x,a}_G}[D(\widetilde{\mathbf{x}},\mathbf{a};\theta_D)] \\ 
&- \lambda \mathbb E_{(\hat{\mathbf{x}},\mathbf{a}) \sim \mathbb P^{x,a}_{\alpha}}[\left(||\nabla_{\hat{\mathbf{x}}}D(\hat{\mathbf{x}},\mathbf{a}; \theta_D)||_2 - 1\right)^2], 
\end{split}
\label{eq:wgan}
\end{equation}

where $\mathbb E[.]$ represents the expected value operator, $\mathbb P_S^{x,a}$ is the joint distribution of visual and semantic features from the seen classes (in practice, samples from that distribution are the ones in $\mathcal{D}_S^{Tr}$), $\mathbb P^{x,a}_G$ represents the joint distribution of semantic features and the visual features produced by the generative model $G(.)$, $\lambda$ denotes the penalty coefficient, and $\mathbb P^{x,a}_{\alpha}$ is the joint distribution of the semantic features and the visual features produced by $\hat{\mathbf{x}} \sim \alpha \mathbf{x} + (1-\alpha)\widetilde{\mathbf{x}}$ with $\alpha \sim \mathcal{U}(0,1)$ (i.e., uniform distribution). 

Finally, the f-CLSWGAN is trained with the following objective function:

\begin{equation}
\theta_G^*,\theta_C^*,\theta_D^*=\arg\min_{\theta_G,\theta_C} \max_{\theta_D} \ell_{WGAN}(\theta_G,\theta_D) + \beta \ell_{CLS}(\theta_C,\theta_G),  
\label{eq:clswgan}
\end{equation}
where $\ell_{CLS}(\theta_C,\theta_G) = -\mathbb E_{(\widetilde{\mathbf{x}},y) \sim \mathbb P^{x,y}_G}[\log P(y | \widetilde{\mathbf{x}}, \theta_C)]$, with 
\begin{equation}
P(y | \widetilde{\mathbf{x}}, \theta_C) = \frac{\exp( (\theta_C(y))^T\widetilde{\mathbf{x}})}{\sum_{c\in \mathcal{Y}}\exp((\theta_C(c))^T\widetilde{\mathbf{x}})}
\label{eq:softmax}
\end{equation}

representing the probability that the sample $\widetilde{\mathbf{x}}$ has been predicted with its true
label $y$, and $\beta$ is a hyper-parameter that weights the contribution of the loss function.  This regularization with the classification loss was found by Xian et al.~\cite{XianCVPR2018} to enforce $G(.)$ to generate discriminative visual representations.  The model obtained from the optimization in (\ref{eq:clswgan}) is referred to as \textbf{baseline} in the experiments.

\subsection{Multi-modal Cycle Consistency Loss}
\label{sec:cycle}

The main issue present in previously proposed GZSL approaches based on generative models~\cite{XianCVPR2018,long2017zero,bucher2017generating} is that the unconstrained nature of the generation process (from semantic to visual features) may produce image representations that are too far from the real distribution present in the training set, resulting in an ineffective multi-class classifier training, particularly for the unseen classes.  The approach we propose to alleviate this problem consists of constraining the synthetic visual representations to generate back their original semantic features - this regularization has been inspired by the cycle consistency loss~\cite{CycleGAN2017}.  Figure~\ref{fig:overview} shows an overview of our proposal.
This approach, representing the main contribution of this paper, is represented by the following loss:

\begin{equation}
\begin{split}
\ell_{CYC}(\theta_R,\theta_G)  & =  \mathbb E_{\mathbf{a} \sim \mathbb P_S^a,\mathbf{z} \sim \mathcal{N}(\mathbf{0},\mathbf{I})} \left [ \|{\mathbf{a} - R(G(\mathbf{a},\mathbf{z};\theta_G);\theta_R)}\|_2^2 \right ]  \\
& + \mathbb E_{\mathbf{a} \sim \mathbb P_U^a,\mathbf{z} \sim \mathcal{N}(\mathbf{0},\mathbf{I})} \left [ \|{\mathbf{a} - R(G(\mathbf{a},\mathbf{z};\theta_G);\theta_R)}\|_2^2 \right ],
\end{split}
\label{eq:cyc}
\end{equation}

where $\mathbb P_S^a$ and $\mathbb P_U^a$ denote the distributions of semantic features of the seen and unseen classes, respectively, and $R:\mathcal{X} \rightarrow \mathcal{A}$ represents a regressor that estimates the original semantic features from the visual representation generated by $G(.)$.

\subsection{Feature Generation}
\label{sec:feature}

Using the losses proposed in Sections~\ref{sec:CLSWGAN} and \ref{sec:cycle}, we can propose several feature generators.  First, we pre-train the regressor $R(.)$ defined below in (\ref{eq:loss_reg}), by minimizing a loss function computed only from the seen classes, as follows:

\begin{equation}
\ell_{REG}(\theta_R) = \mathbb E_{(\mathbf{a},\mathbf{x}) \sim \mathbb P_S^{a,x}} \left [ \|{\mathbf{a} - R(\mathbf{x};\theta_R)}\|_2^2 \right ],
\label{eq:loss_reg}
\end{equation}
where $\mathbb P_S^{a,x}$ represents the real joint distribution of image and semantic features present in the seen classes.  In practice, this regressor is defined by a multi-layer perceptron, whose output activation function  depends on the format of the semantic vector.

Our first strategy to build a feature generator consists of pre-training a regressor (using samples from seen classes) optimized by minimizing $\ell_{REG}$ in (\ref{eq:loss_reg}), which produces $\theta_R^*$ and training the generator and discriminator of the WGAN using the following optimization function:
\begin{equation}
\theta_G^*,\theta_D^* = \arg\min_{\theta_G} \max_{\theta_D} \ell_{WGAN}(\theta_G,\theta_D) + \lambda_1 \ell_{CYC}(\theta_R^*,\theta_G),  
\label{eq:cycle_wgan}
\end{equation}
where $\ell_{WGAN}$ is defined in (\ref{eq:wgan}), $\ell_{CYC}$ is defined in (\ref{eq:cyc}), and $\lambda_1$ weights the importance of the second optimization term.  The optimization in (\ref{eq:cycle_wgan}) can use both the seen and unseen classes, or it can rely only the seen classes, in which case the loss $\ell_{CYC}$ in (\ref{eq:cyc}) has to be modified so that its second term (that depends on unseen classes) is left out of the optimization.  The feature generator model in (\ref{eq:cycle_wgan}) trained with seen and unseen classes is referred to as \textbf{cycle-(U)WGAN}, while the  feature generator trained with only seen classes is labeled \textbf{cycle-WGAN}.

The second strategy explored in this paper to build a feature generator involves pre-training the regressor in (\ref{eq:loss_reg}) using samples from seen classes to produce $\theta_R^*$, and pre-training a softmax classifier for the seen classes using $\ell_{CLS}$, defined in (\ref{eq:clswgan}), which results in $\theta_C^*$.  Then we train the combined loss function:
\begin{equation}
\theta_G^*,\theta_D^* = \arg\min_{\theta_G} \max_{\theta_D} \ell_{WGAN}(\theta_G,\theta_D) + \lambda_1 \ell_{CYC}(\theta_R^*,\theta_G) + \lambda_2\ell_{CLS}(\theta_C^*,\theta_G).  
\label{eq:cycle_clswgan}
\end{equation}
The feature generator model in (\ref{eq:cycle_clswgan}) trained with seen classes is referred to as \textbf{cycle-CLSWGAN}.

\subsection{Learning and Testing}
\label{sec:classifier}

As shown in~\cite{XianCVPR2018} the training of a classifier using a potentially unlimited number of samples from the seen and unseen classes generated with $\mathbf{x} \sim G(\mathbf{a},\mathbf{z};\theta_G^*)$ produces more accurate classification results compared with multi-modal embedding models~\cite{akata2016label,frome2013devise,akata2015evaluation,romera2015embarrassingly}.  Therefore, we train a final softmax classifier $P(y|\mathbf{x},\theta_C)$, defined in (\ref{eq:softmax}), using the generated visual features by minimizing the negative log likelihood loss $\ell_{CLS}(\theta_C,\theta^*_G)$, as defined in (\ref{eq:clswgan}), where $\theta_G^*$ has been learned from one of the feature learning strategies discussed in Sec.~\ref{sec:feature} - the training of the classifier produces $\theta^*_C$. The samples used for training the classifier are generated based on the task to be solved.  For instance, for ZSL, we only use generated visual representations from the set of unseen classes; while for GZSL, we use the generated samples from seen and unseen classes.

Finally, the testing is based on the prediction of a class for an input test visual representation $\mathbf{x}$, as follows:
\begin{equation}
y^* = \arg \max_{y \in \widetilde{\mathcal{Y}}} P(y|\mathbf{x},\theta^*_C),
\end{equation}
where $\widetilde{\mathcal{Y}} = \mathcal{Y}$ for GZSL or $\widetilde{\mathcal{Y}} = \mathcal{Y}_U$ for ZSL.

%
%
%
%
%
%
%
\section{Experiments}
\label{sec:experiments}

In this section, we first introduce the datasets and evaluation criteria used in the experiments, then we  discuss the experimental set-up and finally show the results of our approach, comparing with the state-of-the-art results.

%
%
%
%
\subsection{Datasets} 
\label{sec:datasets}

We evaluate the proposed method on the following ZSL/GZSL benchmark datasets, using the experimental setup of~\cite{XianCVPR2017}, namely: 
CUB-200-2011~\cite{welinder2010caltech,XianCVPR2018},
FLO~\cite{nilsback2008automated},
SUN~\cite{XianCVPR2017}, and
AWA~\cite{lampert2009lerning,XianCVPR2017}  -- where CUB, FLO and SUN are fine-grained datasets, and AWA coarse. Table~\ref{table:dataset-stats} shows some basic information about these datasets in terms of number of seen and unseen classes and number of training and testing images.   
For CUB-200-2011~\cite{welinder2010caltech,XianCVPR2018} and Oxford-Flowers \cite{nilsback2008automated}, the semantic feature has 1024 dimensions produced by the character-based CNN-RNN~\cite{reed2016learning} that encodes the textual description of an image containing fine-grained visual descriptions (10 sentences per image).  
The sentences from the unseen classes are not used for training the CNN-RNN and the per-class sentence is obtained by averaging the CNN-RNN semantic features that belong to
the same class.
For the FLO dataset~\cite{nilsback2008automated}, we used the same type of semantic feature with 1024 dimensions~\cite{reed2016learning} as was used for CUB (please see description above).
For the SUN dataset~\cite{XianCVPR2017}, the semantic features have 102 dimensions.  Following the protocol from Xian et al.\cite{XianCVPR2017}, visual features are represented by the activations of the 2048-dim top-layer pooling units of ResNet-101~\cite{he2016resnet}, obtained from the entire image. 
For AWA~\cite{lampert2009lerning,XianCVPR2017}, we use a semantic feature containing 85 dimensions denoting per-class attributes.
In addition, we also test our approach on \textit{ImageNet}~\cite{deng2009imagenet}, for a split containing 100 classes for testing ~\cite{wang2017multi}.

The input images do not suffer any pre-processing (cropping, background subtraction, etc.) and we do not use any type of
data augmentation.
This ResNet-101 is pre-trained on ImageNet with 1K classes~\cite{deng2009imagenet} and is not fine tuned.
For the synthetic visual representations, we generate 2048-dim CNN features using one of the feature generation models, presented in Sec.~\ref{sec:feature}.

For CUB, FLO, SUN, and AWA we use the zero-shot splits proposed by Xian et al.~\cite{XianCVPR2017}, making sure that none of the training classes are present on ImageNet~\cite{deng2009imagenet}.
Differently from these datasets (i.e., CUB, FLO, SUN, AWA), we observed that there is a lack of standardized experimental setup for GZSL on \textit{Imagenet}. Recently, papers have used \textit{ImageNet} for GZSL using several splits (e.g., 2-hop, 3-hop), but we noticed that some of the supposedly unseen classes can actually be seen during training (e.g., in split \textbf{2-hop}, we note that the class \textit{American mink} is assumed to be unseen, while class \textit{Mink} is seen, but these two classes are arguably the same).  Nevertheless, in order to demonstrate the competitiveness of our proposed \textbf{cycle-WGAN}, we compare it to the \textbf{baseline} using carefully selected 100 unseen classes~\cite{wang2017multi} (i.e., no overlap with 1k training seen classes) from \textit{ImageNet}.

\begin{table}[t]
\centering
\label{table:dataset-stats}
\caption{Information about the datasets CUB\cite{welinder2010caltech}, FLO\cite{nilsback2008automated}, SUN \cite{xiao2010sun}, AWA\cite{XianCVPR2017}, and ImageNet~\cite{deng2009imagenet}. Column (1) shows the number of seen classes, denoted by $|\mathcal{Y}_S|$, split into the number of training and validation classes (train+val), (2) presents the number of unseen classes $| \mathcal{Y}_U |$, (3) displays the number of samples available for training $|\mathcal{D}^{Tr}|$ and (4) shows number of testing samples that belong to the unseen classes $|\mathcal{D}_U^{Te}|$ and number of testing samples that belong to the seen classes $|\mathcal{D}_S^{Te}|$.}
\begin{tabular}{l|c|c|c|c}
\hline
\textbf{Name} & $|\mathcal{Y}_S|$ (train+val) & $|\mathcal{Y}_U|$ & $|\mathcal{D}^{Tr}|$ & $|\mathcal{D}^{Te}_U|+|\mathcal{D}^{Te}_S|$ \\
\hline
CUB     & 150 (100+50) & 50 & 7057 & 1764+2967  \\
FLO     & 82 (62+20) & 20 & 1640 & 1155+5394  \\
SUN    & 745 (580+65) & 72 & 14340 & 2580+1440   \\
AWA    & 40 (27+13) & 10 & 19832 & 4958+5685   \\
ImageNet & 1000 (1000 + 0) & 100 & $1.2 \times 10^6$ & 5200+0    \\
\hline
\end{tabular}
\end{table}
%

%
%
%
%
\subsection{Evaluation Protocol}

We follow the evaluation protocol proposed by Xian et al.~\cite{XianCVPR2017}, where results are based on average per-class top-1 accuracy.  For the ZSL evaluation, top-1 accuracy results are computed with respect to the set of unseen classes $\mathcal{Y}_U$, where the average accuracy is  independently computed for each class, which is then averaged over all unseen classes.  For the GZSL evaluation, we compute the average per-class top-1 accuracy on seen classes $\mathcal{Y}_S$, denoted by $s$, the average per-class top-1 accuracy on unseen classes $\mathcal{Y}_U$, denoted by $u$, and their harmonic mean, i.e. $H = 2 \times (s ∗\times u)/(s + u)$.

%
%
%
%
\subsection{Implementation Details}

In this section, we explain the implementation details of the generator $G(.)$, the discriminator $D(.)$, the regressor $R(.)$, and the weights used for the hyper-parameters in the loss functions in (\ref{eq:wgan}),(\ref{eq:clswgan}),(\ref{eq:cycle_wgan}) and (\ref{eq:cycle_clswgan}) - all these terms have been formally defined in Sec.~\ref{sec:method} and depicted in Fig. ~\ref{fig:overview}.
The generator consists of a multi-layer perceptron (MLP) with a single hidden layer containing 4096 nodes, where this hidden layer is activated by LeakyReLU~\cite{maas2013relu}, and the output layer, with 2048 nodes, has a ReLU activation~\cite{nair2010rectified}.
The weights of $G(.)$ are initialized with a truncated normal initialization with mean 0 and standard deviation $0.01$ and the biases are initialized with $0$. 
The discriminator $D(.)$ is also an MLP consisting of a single hidden layer with 4096 nodes, which is activated by LeakyReLU, and the output layer has no activation.  The initialization of $D(.)$ is the same as for $G(.)$.  
The regressor $R(.)$ is a linear transform from the visual space $\mathcal{X}$ to the semantic space $\mathcal{A}$.
Following~\cite{XianCVPR2018}, we set $\lambda=10$ in (\ref{eq:wgan}), $\beta = 0.01$ in (\ref{eq:clswgan}) and $\lambda_1 = \lambda_2 = 0.01$ in (\ref{eq:cycle_wgan}) and (\ref{eq:cycle_clswgan}). We ran an empirical evaluation with the training set and noticed that when $\lambda_1$ and $\lambda_2$ share the same value, the training becomes stable, but a more systematic evaluation to assess the relative importance of these two hyper-parameters is still needed. Table~\ref{tab:hyp} shows the learning rates for each model (denoted by $lr_{\{ R(.), G(.), D(.)  \}}$), batch sizes (\textbf{batch}) and number of epochs (\textbf{\#ep}) used for each dataset and model -- the values for $G(.)$ and $D(.)$ have been estimated to reproduce the published results of our implementation of f-CLSWGAN (explained below), and the values for $R(.)$ have been estimated by cross validation using the training and validation sets.

Regarding the number of visual representations generated to train the classifier, we performed a few experiments and reached similar conclusions, compared to~\cite{XianCVPR2018}. For all experiments in the paper, we generated 300 visual representations per class~\cite{XianCVPR2018}. We reached this number after a study that shows that for a small number of representations (below 100), the classification results were not competitive; for values superior to 200 or more, results became competitive, but unstable; and above 300, results were competitive and stable.

\begin{table}[t]
\centering
\caption{Summary of cross-validated hyper-parameters in our experiments.}
\label{tab:hyp}
\begin{tabular}{|lcccccccccc|}
\hline
    &  \multicolumn{3}{|c|}{$R(.)$} & \multicolumn{4}{|c|}{GAN: $G(.)$ and $D(.)$} & \multicolumn{3}{|c|}{Classifier} \\
    \hline
    & \multicolumn{1}{|c|}{$lr_{R(.)}$}& \multicolumn{1}{|c|}{\textbf{batch}}    & \multicolumn{1}{|c|}{\textbf{\#ep}}
    & \multicolumn{1}{|c|}{\textbf{$lr_{G(.)}$}}  & \multicolumn{1}{|c|}{\textbf{$lr_{D(.)}$}} & \multicolumn{1}{|c|}{\textbf{batch}} & \multicolumn{1}{|c|}{\textbf{\#ep}} 
    & \multicolumn{1}{|c|}{\textbf{$lr$}}   & \multicolumn{1}{|c|}{\textbf{batch}}  & \multicolumn{1}{|c|}{\textbf{\#ep}} 
    \\ \hline
\multicolumn{1}{|c|}{\textbf{CUB}} & \multicolumn{1}{|c|}{$1e^{-4}$} & \multicolumn{1}{|c|}{$64$} & \multicolumn{1}{|c|}{$100$} & \multicolumn{1}{|c|}{$1e^{-4}$} & \multicolumn{1}{|c|}{$1e^{-3}$} & \multicolumn{1}{|c|}{$64$} & \multicolumn{1}{|c|}{$926$} & \multicolumn{1}{|c|}{$1e^{-4}$} & \multicolumn{1}{|c|}{$4096$} & \multicolumn{1}{|c|}{$80$} \\
\multicolumn{1}{|c|}{\textbf{FLO}} & \multicolumn{1}{|c|}{$1e^{-4}$} & \multicolumn{1}{|c|}{$64$} & \multicolumn{1}{|c|}{$100$} & \multicolumn{1}{|c|}{$1e^{-4}$} & \multicolumn{1}{|c|}{$1e^{-3}$} & \multicolumn{1}{|c|}{$64$} & \multicolumn{1}{|c|}{$926$} & \multicolumn{1}{|c|}{$1e^{-4}$} & \multicolumn{1}{|c|}{$2048$} & \multicolumn{1}{|c|}{$100$} \\
\multicolumn{1}{|c|}{\textbf{SUN}} & \multicolumn{1}{|c|}{$1e^{-4}$} & \multicolumn{1}{|c|}{$64$} & \multicolumn{1}{|c|}{$100$} & \multicolumn{1}{|c|}{$1e^{-2}$} & \multicolumn{1}{|c|}{$1e^{-2}$} & \multicolumn{1}{|c|}{$64$} & \multicolumn{1}{|c|}{$926$} & \multicolumn{1}{|c|}{$1e^{-4}$} & \multicolumn{1}{|c|}{$4096$} & \multicolumn{1}{|c|}{$298$} \\
\multicolumn{1}{|c|}{\textbf{AWA}} & \multicolumn{1}{|c|}{$1e^{-3}$} & \multicolumn{1}{|c|}{$64$} & \multicolumn{1}{|c|}{$50$} & \multicolumn{1}{|c|}{$1e^{-4}$} & \multicolumn{1}{|c|}{$1e^{-3}$} & \multicolumn{1}{|c|}{$64$} & \multicolumn{1}{|c|}{$350$} & \multicolumn{1}{|c|}{$1e^{-4}$} & \multicolumn{1}{|c|}{$2048$} & \multicolumn{1}{|c|}{$37$} \\

\multicolumn{1}{|c|}{\textit{ImageNet}} & \multicolumn{1}{|c|}{$1e^{-4}$} & \multicolumn{1}{|c|}{$2048$} & \multicolumn{1}{|c|}{$5$} & \multicolumn{1}{|c|}{$1e^{-4}$} & \multicolumn{1}{|c|}{$1e^{-3}$} & \multicolumn{1}{|c|}{$256$} & \multicolumn{1}{|c|}{$300$} & \multicolumn{1}{|c|}{$1e^{-3}$} & \multicolumn{1}{|c|}{$2048$} & \multicolumn{1}{|c|}{$300$}

\\ \hline
\end{tabular}
\end{table}

\begin{table}[t]
\caption{Comparison between the  reported results of \textbf{f-CLSWGAN}~\cite{XianCVPR2018} and our implementation of it, labeled \textbf{baseline}, 
where we show the top-1 accuracy on the unseen test $\mathcal{Y}_U$ (GZSL), the top-1 accuracy for seen test $\mathcal{Y}_S$ (GZSL), the harmonic mean $H$ (GZSL), and the top-1 accuracy for ZSL ($T1_Z$).}
\label{table:lit_baseline}

\centering

\resizebox{\textwidth}{!}{
\begin{tabularx}{\textwidth}{l|cccc|cccc|cccc|cccc}
\hline
& &  \scriptsize\shortstack{\textbf{CUB}} & &
& &  \scriptsize\shortstack{\textbf{FLO}} & &
& &  \scriptsize\shortstack{\textbf{SUN}} & &
& &  \scriptsize\shortstack{\textbf{AWA}} & & 
\\
 \scriptsize\shortstack{\textbf{Classifier}} 
&   \scriptsize\shortstack{$\mathcal{Y}_U$} &  \scriptsize\shortstack{$\mathcal{Y}_S$} &  \scriptsize\shortstack{$H$} &  \scriptsize\shortstack{$T1_Z$}
&  \scriptsize\shortstack{$\mathcal{Y}_U$} & \scriptsize\shortstack{$\mathcal{Y}_S$} & \scriptsize\shortstack{$H$} & \scriptsize\shortstack{$T1_Z$}
&  \scriptsize\shortstack{$\mathcal{Y}_U$} & \scriptsize\shortstack{$\mathcal{Y}_S$} & \scriptsize\shortstack{$H$} & \scriptsize\shortstack{$T1_Z$}
&  \scriptsize\shortstack{$\mathcal{Y}_U$} & \scriptsize\shortstack{$\mathcal{Y}_S$} & \scriptsize\shortstack{$H$} & \scriptsize\shortstack{$T1_Z$}
\\ \hline
\hline
\tiny\shortstack{\textbf{f-CLSWGAN}}\cite{XianCVPR2018}
& \scriptsize\shortstack{$43.7$} & \scriptsize\shortstack{$57.7$} & \scriptsize\shortstack{$49.7$} & \scriptsize\shortstack{$57.3$}
& \scriptsize\shortstack{$59.0$} & \scriptsize\shortstack{$73.8$} & \scriptsize\shortstack{$65.6$} & \scriptsize\shortstack{$67.2$}
& \scriptsize\shortstack{$42.6$} & \scriptsize\shortstack{$36.6$} & \scriptsize\shortstack{$39.4$} & \scriptsize\shortstack{$60.8$}
& \scriptsize\shortstack{$57.9$}  & \scriptsize\shortstack{$61.4$} & \scriptsize\shortstack{$59.6$}  & \scriptsize\shortstack{$68.2$}
\\
\scriptsize\shortstack{\textbf{baseline}}
& \scriptsize\shortstack{$43.8$} & \scriptsize\shortstack{$60.6$} & \scriptsize\shortstack{$50.8$} & \scriptsize\shortstack{$57.7$}
& \scriptsize\shortstack{$58.8$} & \scriptsize\shortstack{$70.0$} & \scriptsize\shortstack{$63.9$} & \scriptsize\shortstack{$66.8$}
& \scriptsize\shortstack{$47.9$} & \scriptsize\shortstack{$32.4$} & \scriptsize\shortstack{$38.7$} & \scriptsize\shortstack{$58.5$}
& \scriptsize\shortstack{$56.0$}  & \scriptsize\shortstack{$62.8$} & \scriptsize\shortstack{$59.2$}  & \scriptsize\shortstack{$64.1$}
\\ \hline
\end{tabularx}
}
\end{table}

%
%
%
%
Since our approach is based on the f-CLSWGAN~\cite{XianCVPR2018}, we re-implemented this methodology.
In the experiments, the results from our implementation of f-CLSWGAN using a softmax classifier is labeled as \textbf{baseline}.
The results that we obtained from our baseline are very similar to the reported results in~\cite{XianCVPR2018}, as shown in Table ~\ref{table:lit_baseline}. 
For ImageNet, note that we use a split~\cite{wang2017multi} that is different from previous ones used in the literature, as explained above in Sec.~\ref{sec:datasets}, so it is not possible to have a direct comparison between f-CLSWGAN~\cite{XianCVPR2018} and our \textbf{baseline}.  Nevertheless, we show in Table~\ref{table:zsl_imagenet} that the results we obtain for the split~\cite{wang2017multi} are in fact similar to the reported results  for f-CLSWGAN~\cite{XianCVPR2018} for similar ImageNet splits.
We developed our code~\footnote{Code is available at: https://github.com/rfelixmg/frwgan-eccv18} and perform all experiments using Tensorflow~\cite{abadi2016tensorflow}.

%
%
%
%
%
%
%
%
%
\section{Results}

In this section we show the GZSL and ZSL results using our proposed models \textbf{cycle-WGAN}, \textbf{cycle-(U)WGAN}  and \textbf{cycle-CLSWGAN}, the baseline model f-CLSWGAN, denoted by \textbf{baseline}, and several other baseline methods previously used in the field for benchmarking~\cite{XianCVPR2017}. 
Table~\ref{table:gzsl_results} shows the \textbf{GZSL results} and Table~\ref{table:zsl_results} shows the \textbf{ZSL results} obtained from our proposed methods, and several baseline approaches on CUB, FLO, SUN and AWA datasets.
The results in Table~\ref{table:zsl_imagenet} shows that the top-1 accuracy on ImageNet for \textbf{cycle-WGAN} and \textbf{baseline}~\cite{XianCVPR2018}.

\input{src/gzsl_table.tex}

\input{src/zsl_table.tex}

\begin{table}
\centering
\caption{ZSL and GZSL ImageNet results using per-class average top-1 accuracy on the test sets of unseen classes $\mathcal{Y}_U$ -- all results shown in percentage.}
\label{table:zsl_imagenet}
\centering
\begin{tabular}{lcc}
\hline
\textbf{Classifier}  & \textbf{ZSL} & \textbf{GZSL}
\\ 
\hline
\textbf{baseline}~\cite{XianCVPR2018} & 7.5 & 0.7 \\
\texttt{cycle-WGAN}  & 8.7 & 1.5\\
\hline
\end{tabular}
\end{table}

%
%
%
%
%

\section{Discussion}
\label{sec:discussion}

Regarding the GZSL results in Table~\ref{table:gzsl_results}, we notice that there is a clear trend of all of our proposed feature generation methods (\textbf{cycle-WGAN}, \textbf{cycle-(U)WGAN}), and \textbf{cycle-CLSWGAN}) to perform better than \textbf{baseline} on the unseen test set.  In particular, it seems advantageous to use the synthetic samples from unseen classes to train the \textbf{cycle-(U)WGAN} model since it achieves the best top-1 accuracy results in 3 out of the 4 datasets, with improvements from 0.7\% to more than 4\%.  In general, the top-1 accuracy improvement achieved by our approaches in the seen test set is less remarkable, which is expected given that we prioritize to improve the results for the unseen classes.  Nevertheless, our approaches achieved improvements from 0.4\% to more than 2.5\% for the seen classes.  Finally, the harmonic mean results also show that our approaches improve over the \textbf{baseline} in a range of between 1\% and 2.2\%.  Notice that this results are remarkable considering the outstanding improvements achieved by f-CLSWGAN~\cite{XianCVPR2018}, represented here by \textbf{baseline}.  In fact, our proposed methods produce the current state of the art GZSL results for these four datasets.

Analyzing the ZSL results in Table~\ref{table:zsl_results}, we again notice that, similarly to the GZSL case, there is a clear advantage in using the synthetic samples from unseen classes to train the \textbf{cycle-(U)WGAN} model.  For instance, top-1 accuracy results show that we can improve over the \textbf{baseline} from 0.9\% to 3.5\%. The results  in this table show that our proposed approaches currently hold the best ZSL results for these datasets.

It is interesting to see that, compared to GZSL, the ZSL results from previous method in the literature are far more competitive, achieving results that are relatively close to ours and the \textbf{baseline}.  
This performance gap between ZSL and GZSL, shown by previous methods, enforces the argument in favor of using generative models to synthesize images from seen and unseen classes to train GZSL models~\cite{XianCVPR2018,bucher2017generating,long2017zero}. As argued throughout this paper, the performance produced by generative models can be improved further with methods that help the training of GANs, such as the cycle consistency loss~\cite{CycleGAN2017}. 

In fact, the experiments clearly demonstrate the advantage of using our proposed multi-modal cycle consistency loss in training GANs for GZSL and ZSL.  In particular, it is interesting to see that the use of synthetic examples of unseen classes generated by \textbf{cycle-(U)WGAN} to train the GZSL classifier provides remarkable improvements over the \textbf{baseline}, represented by f-CLSWGAN~\cite{XianCVPR2018}.  The only exception is with the SUN dataset, where the best result is achieved by \textbf{cycle-CLSWGAN}.  We believe that \textbf{cycle-(U)WGAN} is not the top performer on SUN due to the number of classes and the proportion of seen/unseen classes in this dataset. For CUB, FLO and AWA we notice that there is roughly a $(80\%,20\%)$ ratio between seen and unseen classes.  In contrast, SUN has a $(91\%,9\%)$ ratio between seen and unseen classes. We also notice a sharp increase in the number of classes from 50 to 817 -- GAN models tend not to work well with such a large number of classes. 
Given the wide variety of GZSL datasets available in the field, with different number of classes and seen/unseen proportions, we believe that there is still lots of room for improvement for GZSL models.

Regarding the large-scale study on ImageNet, the results in Table~\ref{table:zsl_imagenet} show that the top-1 accuracy classification
results for \textbf{Baseline} and \textbf{cycle-WGAN} are quite low (similarly to the results observed in~\cite{XianCVPR2018} for several ImageNet splits),
but our proposed approach still shows more accurate ZSL and GZSL classification.

\begin{figure}[t]
\centering
\includegraphics[width=\textwidth]{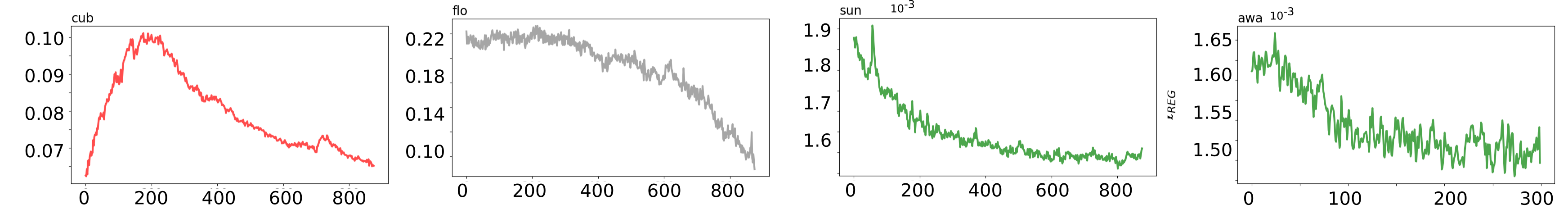}
\caption{Evolution of $\ell_{REG}$ in terms of the number of epochs for CUB, FLO, SUN and AWA.}
\label{fig:sun_regressor}
\end{figure}

\begin{figure}[t]
\centering
	\includegraphics[width=\textwidth]{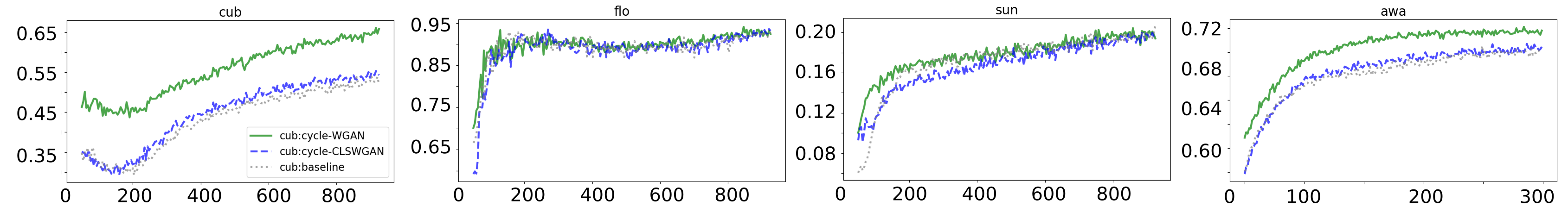}
 	\caption{Convergence of the top-1 accuracy in terms of the number of epochs for the generated training samples from the seen classes for CUB, FLO, SUN and AWA.}
    \label{fig:fake_acc}
\end{figure}

An important question about out approach is whether the regularisation succeeds in mapping the generated visual representations back to the semantic space.  In order to answer this question, we show in Fig.~\ref{fig:sun_regressor} the evolution of the reconstruction loss $\ell_{REG}$ in \eqref{eq:loss_reg} as a function of the number of epochs.  In general, the reconstruction loss decreases steadily over training, showing that our model succeeds at such mapping.
Another relevant question is if our proposed methods take more or less epochs to converge, compared to the \textbf{Baseline} -- Fig.~\ref{fig:fake_acc} shows the classification accuracy of the generated training samples from the seen classes for the proposed models \textbf{cycle-WGAN} and \textbf{cycle-CLSWGAN}, and also for the \textbf{baseline} (note that \textbf{cycle-(U)WGAN} is a fine-tuned model from the \textbf{cycle-WGAN}, so their loss functions are in fact identical for the seen classes shown in the graph).  For three out of four datasets, our proposed \textbf{cycle-WGAN} converges faster. However, when the $\ell_{CLS}$ in included in \eqref{eq:cycle_wgan} to form the loss in \eqref{eq:cycle_clswgan} (transforming \textbf{cycle-WGAN} into \textbf{cycle-CLSWGAN}), then the convergence of \textbf{cycle-CLSWGAN} is comparable to that of the \textbf{baseline}.  Hence, \textbf{cycle-WGAN} tends to converge faster than the \textbf{baseline} and \textbf{cycle-CLSWGAN}.

%
%
%
%
%
%
%

\section{Conclusions and Future Work}

In this paper, we propose a new method to regularize the training of GANs in GZSL models.  The main argument explored in the paper is that the use of GANs to generate seen and unseen synthetic examples for training GZSL models has shown clear advantages over previous approaches.  However, the unconstrained nature of the generation of samples from unseen classes can produce models that may not work robustly for some unseen classes.  Therefore, by constraining the generation of samples from unseen classes, we target to improve the GZSL classification accuracy.  Our proposed constraint is motivated by the cycle consistency loss~\cite{CycleGAN2017}, where we enforce that the generated visual representations maps back to their original semantic feature -- this represents the multi-modal cycle consistency loss.
Experiments show that the use of such loss is clearly advantageous, providing improvements over the current state of the art f-CLSWGAN~\cite{XianCVPR2018} both in terms of GZSL and ZSL.

As noticed in Sec.~\ref{sec:discussion}, GAN-based GZSL approaches offer indisputable advantage over previously proposed methods.  However, the reliance on GANs to generate samples from unseen classes is challenging because GANs are notoriously difficult to train, particularly in unconstrained and large scale problems.  Therefore, future work in this field should be focused on targeting these problems.  In this paper, we provide a solution that addresses the unconstrained problem, but it is clear that other regularization approaches could also be used.  In addition, the use of GANs in large scale problems (regarding the number of classes) should also be more intensively studied, particularly when dealing with real-life datasets and scenarios.  Therefore, we will focus our future research activities in solving these two issues in GZSL.

%
%
%
%
%
%
%
\bibliographystyle{splncs}
\bibliography{egbib}
\end{document}

%% file: src/gzsl_table.tex
\begin{table}[ht]
\centering
\caption{GZSL results using per-class average top-1 accuracy on the test sets of unseen classes $\mathcal{Y}_U$, seen classes $\mathcal{Y}_S$, and the harmonic mean result $H$ -- all results shown in percentage. Results from previously proposed methods in the field extracted from~\cite{XianCVPR2017}.}
\label{table:gzsl_results}
\centering
\resizebox{\textwidth}{!}{
\begin{tabular}{|l|lll|lll|lll|lll|}
\hline
& & \textbf{CUB} & 
& & \textbf{FLO} & 
& & \textbf{SUN} & 
& & \textbf{AWA} & 
\\
\textbf{Classifier}  
&  $\mathcal{Y}_U$ & $\mathcal{Y}_S$ & $H$
&  $\mathcal{Y}_U$ & $\mathcal{Y}_S$ & $H$
&  $\mathcal{Y}_U$ & $\mathcal{Y}_S$ & $H$
&  $\mathcal{Y}_U$ & $\mathcal{Y}_S$ & $H$
\\ \hline

DAP~\cite{lampert2009lerning}  & $ 4.2$ & $25.1$ & $ 7.2$ & $ -  $ & $ -  $ & $  - $ & $ 1.7$ & $67.9$ & $ 3.3$ & $ 0.0$ & $\textbf{88.7}$ & $ 0.0$ \\
IAP~\cite{lampert2009lerning}  & $ 1.0$ & $37.8$ & $ 1.8$ & $ -  $ & $ -  $ & $  - $ & $ 0.2$ & $\textbf{72.8}$ & $ 0.4$ & $ 2.1$ & $78.2$ & $ 4.1$ \\
DEVISE~\cite{frome2013devise}     & $23.8$ & $53.0$ & $32.8$ & $ 9.9$ & $44.2$ & $16.2$ & $16.9$ & $27.4$ & $20.9$ & $13.4$ & $68.7$ & $22.4$ \\
SJE~\cite{akata2015evaluation}        & $23.5$ & $59.2$ & $33.6$ & $13.9$ & $47.6$ & $21.5$ & $14.7$ & $30.5$ & $19.8$ & $11.3$ & $74.6$ & $19.6$\\
LATEM~\cite{xian2016latent}      & $15.2$ & $57.3$ & $24.0$ & $ 6.6$ & $47.6$ & $11.5$ & $14.7$ & $28.8$ & $19.5$ & $ 7.3$ & $71.7$ & $13.3$ \\
ESZSL~\cite{romera2015embarrassingly}      & $12.6$ & $\textbf{63.8}$ & $21.0$ & $11.4$ & $56.8$ & $19.0$ & $11.0$ & $27.9$ & $15.8$ & $ 6.6$ & $75.6$ & $12.1$ \\
ALE~\cite{akata2016label}        & $23.7$ & $62.8$ & $34.4$ & $13.3$ & $61.6$ & $21.9$ & $21.8$ & $33.1$ & $26.3$ & $16.8$ & $76.1$ & $27.5$ \\
SAE~\cite{kodirov2017semantic}        & $ 8.8$ & $18.0$ & $11.8$ & $ -  $ & $ -  $ & $  - $ & $ 7.8$ & $54.0$ & $13.6$ & $ 1.8$ & $77.1$ & $ 3.5$ \\

\hline
\textbf{baseline} \cite{XianCVPR2018} 
& $43.8$ & $60.6$ & $50.8$ 
& $58.8$ & $70.0$ & $63.9$
& $47.9$ & $32.4$ & $38.7$
& $56.0$  & $62.8$ & $59.2$
\\

\texttt{cycle-WGAN} 
& $46.0$ & $60.3$ & $52.2$ 
& $59.1$ & $71.1$ & $64.5$ 
& $48.3$ & $33.1$ & $39.2$ 
& $56.4$ & $63.5$ & $59.7$ 
\\ 

\texttt{cycle-CLSWGAN} 
& $45.7$ & $61.0$ & $52.3$ 
& $59.2$ & $\textbf{72.5}$ & $65.1$ 
& $\textbf{49.4}$ & $33.6$ & $\mathbf{40.0}$
&$56.9$ & $64.0$ & $\textbf{60.2}$
 \\ 
 
\texttt{cycle-(U)WGAN} 
& $\textbf{47.9}$ & $59.3$ & $\mathbf{53.0}$ 
& $\textbf{61.6}$ & $69.2$ & $\textbf{65.2}$ 
& $47.2$ & $33.8$ & $39.4$ 
& $\textbf{59.6}$ & $63.4$ & $59.8$
 \\ \hline
 
              \end{tabular}
}
\end{table}

%% file: src/zsl_table.tex
\begin{table}
\centering
\caption{ZSL results using per-class average top-1 accuracy on the test set of unseen classes $\mathcal{Y}_U$ -- all results shown in percentage. Results from previously proposed methods in the field extracted from~\cite{XianCVPR2017}.}
\label{table:zsl_results}
\centering
\begin{tabular}{l|c|c|c|c}
\hline
& & \textbf{ZSL} &
\\
\textbf{Classifier}  
& \textbf{CUB} 
& \textbf{FLO} 
& \textbf{SUN}
& \textbf{AWA }
\\ \hline
DEVISE~\cite{frome2013devise}          & $52.0$ & $45.9$ & $56.5$ & $54.2$ \\
SJE~\cite{akata2015evaluation}         & $53.9$ & $53.4$ & $53.7$ & $65.6$ \\
LATEM~\cite{xian2016latent}            & $49.3$ & $40.4$ & $55.3$ & $55.1$ \\
ESZSL~\cite{romera2015embarrassingly}  & $53.9$ & $51.0$ & $54.5$ & $58.2$ \\
ALE~\cite{akata2016label}              & $54.9$ & $48.5$ & $58.1$ & $59.9$
\\ \hline

\textbf{baseline} \cite{XianCVPR2018} 

& $57.7$ & $66.8$ & $58.5$ & $64.1$
\\

\texttt{cycle-WGAN} 
& $57.8$ & $68.6$ & $59.7$ & $65.6$
\\

\texttt{cycle-CLSWGAN} 
& $58.4$ & $70.1$ & $\textbf{60.0}$ & $66.3$
 \\
 
\texttt{cycle-(U)WGAN} 
& $\mathbf{58.6}$ & $\textbf{70.3}$  & $59.9$ & $\textbf{66.8}$
 \\ \hline
 
 \end{tabular}
\end{table}